\documentclass[conference]{IEEEtran}
\IEEEoverridecommandlockouts
\usepackage{cite}
\usepackage{amsmath,amssymb,amsfonts}
\usepackage{algorithmic}
\usepackage{graphicx}
\usepackage{float}
\usepackage{textcomp}
\usepackage{xcolor}
\usepackage{subfigure}
\usepackage{multirow}
\usepackage{comment}
\usepackage{url}
\usepackage{booktabs,ragged2e}
\usepackage[mathlines,switch]{lineno}
\DeclareMathOperator*{\argminA}{arg\,min}

\begin{document}

\title{
{\footnotesize This is an archival version of this paper. Please cite the published version DOI: \url{https://doi.org/10.1109/MN55117.2022.9887703}}\\
Development of a Cooperative Localization System \\using a UWB Network and BLE Technology
}

\author{\IEEEauthorblockN{Valerio Brunacci, Alessio De Angelis, Gabriele Costante}
\IEEEauthorblockA{\textit{Department of Engineering }\\
\textit{University of Perugia}\\
Perugia, Italy\\
valerio.brunacci@studenti.unipg.it, alessio.deangelis@unipg.it, gabriele.costante@unipg.it}
}

\maketitle

\begin{abstract}
This paper presents the development of a system able to estimate the 2D relative position of nodes in a wireless network, based on distance measurements between the nodes.
The system uses ultra wide band ranging technology and the Bluetooth Low Energy protocol to acquire data. 
Furthermore, a nonlinear least squares problem is formulated and solved numerically for estimating the relative positions of the nodes. 
The localization performance of the system is validated by experimental tests, demonstrating the capability of measuring the relative position of a network comprised of 4 nodes with an accuracy of the order of 3 cm and an update rate of 10 Hz. 
This shows the feasibility of applying the proposed system for multi-robot cooperative localization and formation control scenarios.
\end{abstract}


\section{Introduction}
The task of localization is now common to various fields such as mobile robotics, the Internet of Things, Smart Cities and Smart Industries.
Therefore, developing low-cost, low-power devices that can locate themselves is a key factor for successful applications in these fields.
The most popular localization systems are based on satellite communications, i.e. GNSS (Global Navigation Satellite System).
However, GNSSs do not perform as well indoors or where the direct visibility of satellites is compromised, for example by buildings or trees \cite{7762095}.
In the latter type of situation, research is very active in finding a new reference system.
Many approaches exist: ranging from systems based on existing wireless networks, using vision with a known map or exploiting magnetic systems with coils that achieve millimeter precision \cite{7880609}.

In this work, a system able to perform cooperative localization based on Ultra Wide Band (UWB) and Bluetooth Low Energy (BLE) sensors is presented.
One of the key contributions of this research is the development of a fully-wireless, inexpensive and cooperative system for relative localization.
Other works, in fact, such as \cite{9217573} and \cite{Jimnez2021ImprovingTA},  exploit wired connections for reading UWB device data and therefore do not have a fully-wireless setup. 
The long-term goal of this system is to provide real-time relative positions for dynamic robotic systems.

The proposed system consists of 4 nodes and 5 Decawave DW1001 UWB transceivers capable of distance estimation with other devices using Two-Way-Ranging (TWR).
These devices are not designed to perform cooperative measurements. Instead, they have been conceived and realized for an Anchor-Tag scenario. In this scenario, static and fixed nodes (Anchors) provide distance measurements towards one or more tags that are free to move in the anchor coverage area.
In \cite{Jimnez2021ImprovingTA}, an extensive analysis of the Decawave system is described, aiming at improving its performance and scalability.
Anchor-Tag measurements constitute the input for the localization algorithm present in the factory firmware.

This localization algorithm solves a trilateration problem similar to that solved by GNSS.
Its solution is based on two key aspects: the definition of a cost function and its minimization.
In order to solve that mathematical problem, it is possible to range from linear and non-linear least-squares formulations, or to more advanced statistical analysis techniques such as Multidimensional Scaling (MDS) \cite{MDS}.

\begin{figure}[t]
    \centerline{\includegraphics[width=0.4\columnwidth]{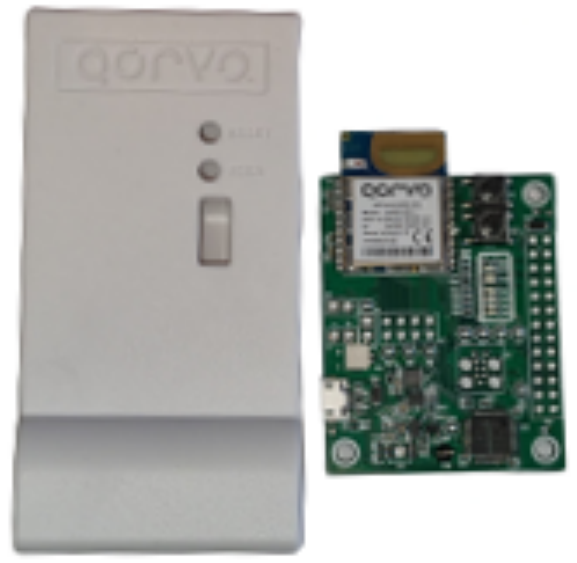}}
    \caption{Photo of a node of the proposed system, consisting of a DWM1001-DEV development board (right) with its casing (left).}
    \label{fig:device}
\end{figure}

\section{Methodology}
In this section, the implementation of the proposed cooperative localization system is described, together with the main software and hardware design choices, and the employed conventions.

\subsection{Decawave DWM1001 System Analysis}
\label{s:decawave}
The proposed localization system is comprised of several nodes, as shown in Fig. \ref{fig:device}, forming a UWB network. Each of the nodes is based on a Decawave DWM1001 UWB radio module, which is capable of a nominal ranging accuracy of approximately 10 cm \cite{decawaveDWM1001}.
This module is also equipped with a BLE interface.
In particular, to implement each node, a DWM1001-DEV development board is used, which contains a DWM1001 UWB module together with a micro-usb interface, a microcontroller, and additional components.

The manufacturer provides three types of interface between the UWB network and an external control network \cite{decawave}. 
Specifically, it is possible to connect a PC to a node of the network using the UART or SPI serial protocols via the usb connector.
Additionally, a wireless connection may be established by using an external BLE module or device, such as a smartphone or a PC. 
Finally, for larger-coverage configurations, the usage of a gateway is suggested, in order to connect different UWB sub-networks.

Within the UWB network, there are two types of nodes: anchors and tags. 
The anchors are known-position nodes, also known as beacons, whereas the tags are nodes whose positions are estimated by the system.
Distance measurements are performed only between nodes of different types, i.e. only between an anchor and a tag, with a maximum update rate of 10 Hz. 
This is also the maximum update rate of the position estimates provided by the Decawave software. 
\begin{figure}[t]
    \centerline{\includegraphics[width=0.9\columnwidth]{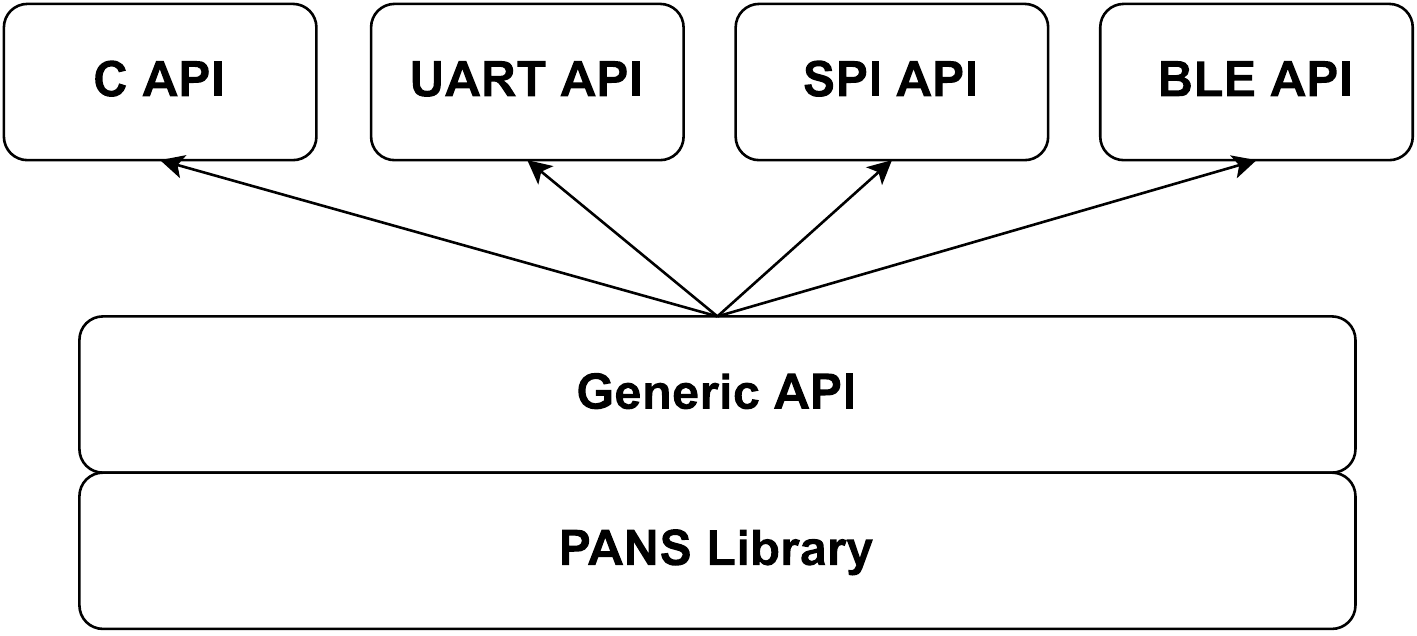}}
    \caption{DWM1001 API Architecture.}
    \label{fig:DWM1001_FirmwareArchitecture}
\end{figure}
The devices are equipped with Application Programming Interfaces (APIs) for all the different communication channels, sharing the functionalities provided by the same basis layer, as shown in Fig.  \ref{fig:DWM1001_FirmwareArchitecture}.
As illustrated by this diagram, the PANS (Positioning and Networking Stack) firmware library is located below the APIs.
This firmware library is developed by Decawave and defines all interactions of the nodes in the UWB network. 
The ranging system implemented by PANS is based on a Time Division Multiple Access mechanism and applies the double-sided TWR technique, between anchor and tag to measure distance \cite{7822844}.

\begin{figure}[t]
    \centerline{\includegraphics[width=0.9\columnwidth]{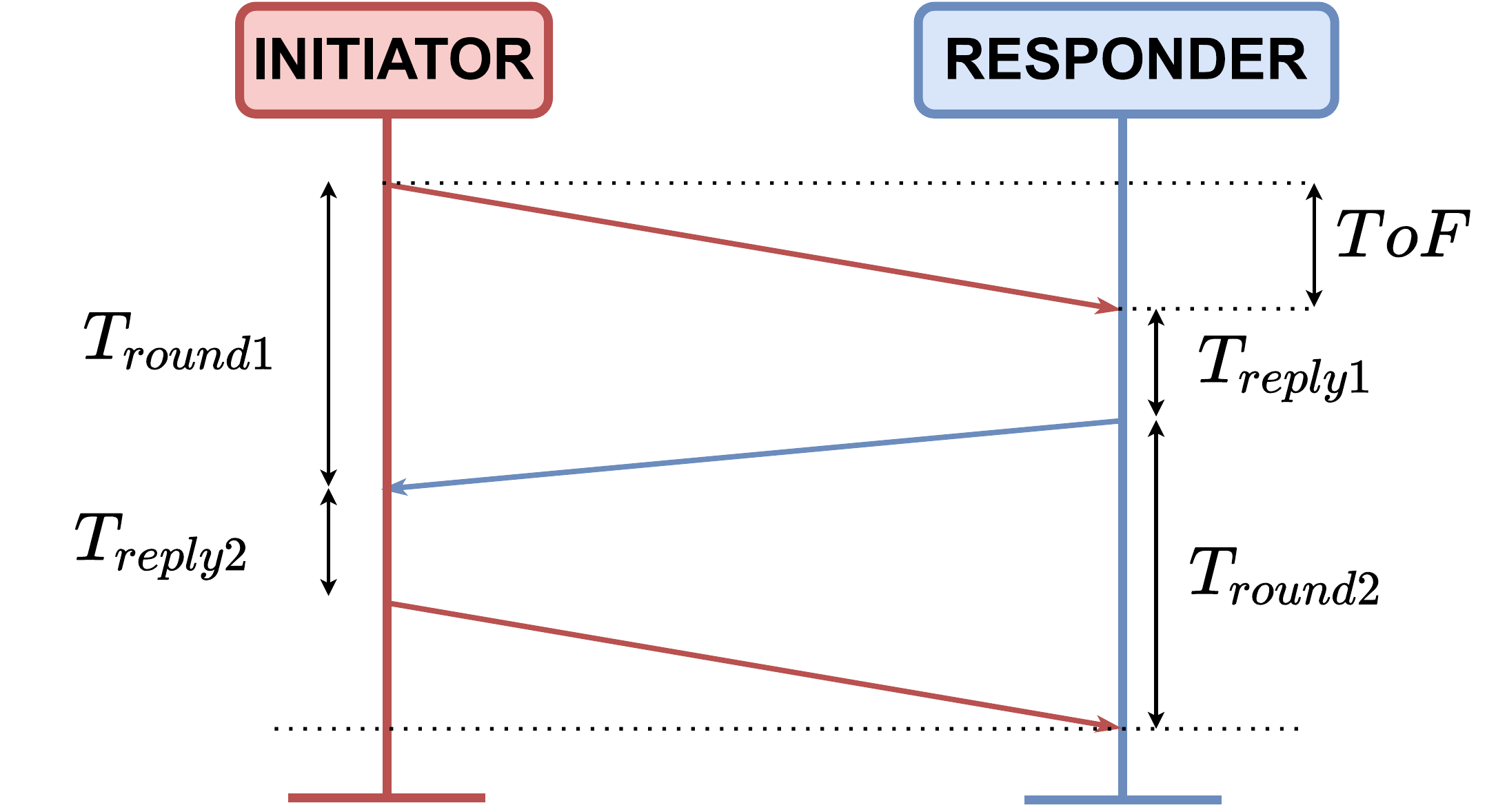}}
    \caption{Implementation of the ToF measurement mechanism.}
    \label{fig:Tof}
\end{figure}

In TWR the communication is established between two devices: an initiator and a responder. The initiator sends a packet, then the responder replies and finally, after the initiator receives that reply, it sends a final message. 
The distance is estimated by measuring the time–of–flight (ToF), defined as follows:
\begin{equation}
ToF=\tfrac{T_{round1} \ast T_{round2}- T_{reply1} \ast T_{reply2}}
{T_{round1} + T_{round2} + T_{reply1} + T_{reply2}}
\label{eq:TOF}
\end{equation}
where the symbols are defined as in Figure \ref{fig:Tof} and the derivation is provided in \cite{7822844}.

\subsection{Proposed System}
\label{section:Proposed System}
The goal of the proposed system is to perform relative cooperative positioning in two dimensions, based on measurements of the distance between the nodes. 
This requires a sufficient number of measurements so that the relative position of each node of the network can be estimated by trilateration. 

To reach this goal, a limitation is given by the PANS firmware, in which distance measurements can only be performed between anchors and tags, as mentioned in Section \ref{s:decawave}. 
Notice that this is just a software limitation and some solutions have been proposed in the literature to circumvent it, such as the system presented in \cite{9341042}.
However, these solutions require a firmware modification, thus present some practical difficulties.
For this reason, a different solution that does not require firmware modifications is employed in this paper, based on a specific network configuration. 

\begin{figure}[t]
    \centerline{\includegraphics[width=0.9\columnwidth]{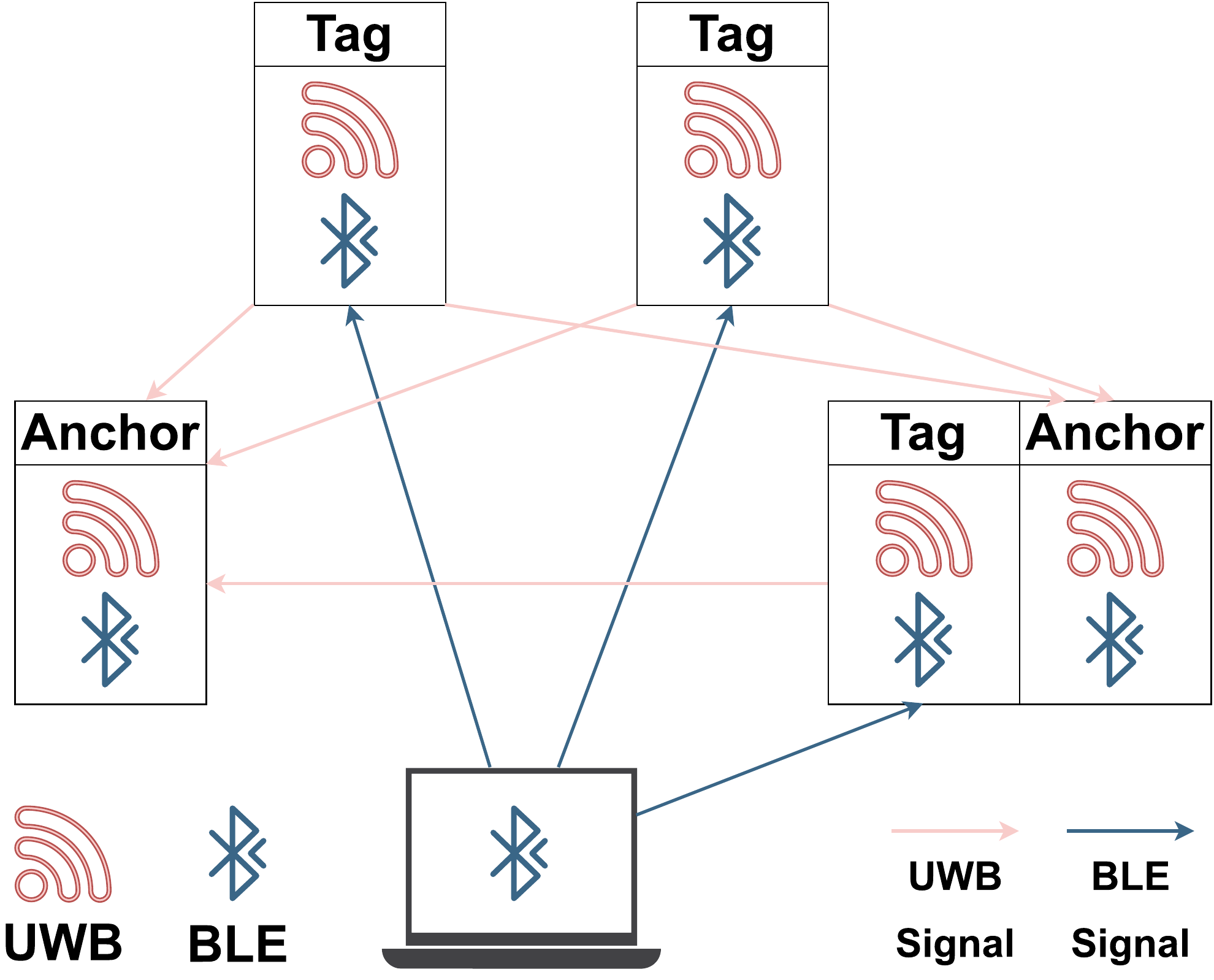}}
    % \caption{Communication scheme of node UWB Network and BLE connection from an external device}
    \caption{Diagram of the UWB connections between nodes and BLE connections to an external device (PC).}
    \label{fig:Node_Configuration}
\end{figure}

In particular, the UWB nodes are placed at the corners of a quadrilateral, properly alternating anchors and tags to ensure that each node is involved in at least two distance measurements. 
Moreover, one of the nodes is comprised of two devices: an anchor and a tag. 
This allows to obtain all required distance measurements.
Finally, to access the information from the UWB network, i.e. the distance measurement results, an external PC establishes a BLE connection with each tag node.
The complete network connection scheme is shown in Fig. \ref{fig:Node_Configuration}.
In the following subsections, the software architecture, the conventions adopted, and the localization method are described in detail. 

\subsubsection{Software Architecture}
\label{s:software}
\begin{figure}[t]
    \centerline{\includegraphics[width=0.9\columnwidth]{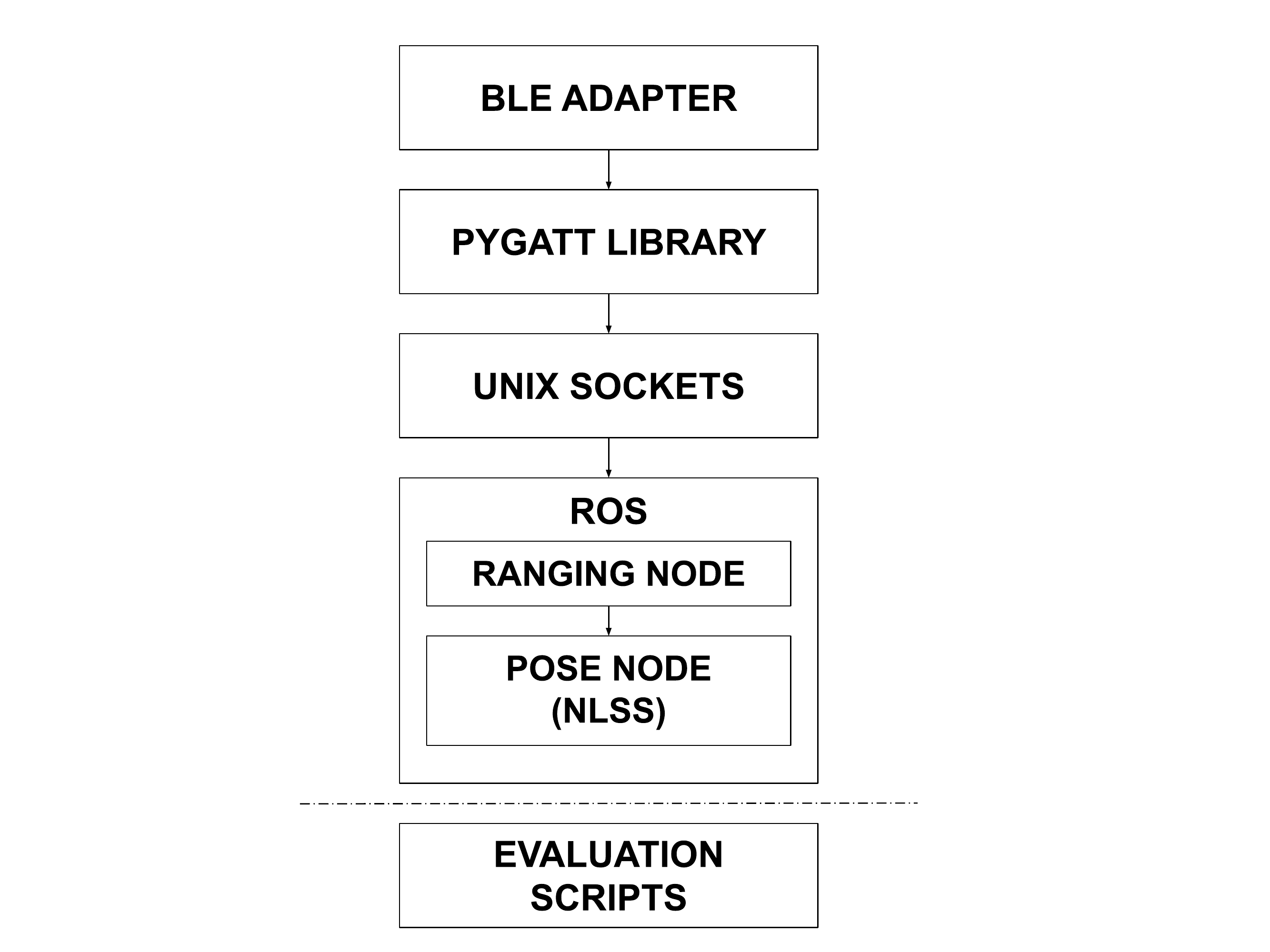}}
    \caption{Software architecture of the proposed system.}
    \label{fig:ProposedSolutionSwStack}
\end{figure}

The main components of the implemented software architecture, as shown in Fig. \ref{fig:ProposedSolutionSwStack}, are the ROS (Robot Operating System) framework, the python PyGatt library \cite{PygattGithub}, and a Non-Linear Least Squares (NLSS) algorithm.

The software is executed on an external PC connected via BLE to the UWB tags. 
The measurement data received by the Bluetooth adapter in the PC are acquired thanks to the PyGatt library, a python wrapper that allows for reading and writing GATT (Generic Attribute Profile) descriptors. 
In this case, PyGatt is used to subscribe to the "Location data" characteristic of the tag nodes. Using this characteristic with notifications enabled, distance measurements can be read.

After being decoded and organized by means of custom structures, the data are put into distinct sockets, one for each tag. 
The sockets are then read by a custom ROS node, denoted as "Ranging Node" in Fig. \ref{fig:ProposedSolutionSwStack}, which writes them into specific topics. 
The second ROS node, denoted as "Pose Node", computes an estimate of the positions of all nodes using range measurements acquired from the ROS topics mentioned above. 
The Pose Node uses the "trf" (Trust Region Reflective) NLSS algorithm, provided by the python \emph{scipy} library \cite{2020SciPy-NMeth}, to estimate the position numerically  by minimizing the cost function that will be described in Section \ref{s:localization_method}. 

\subsubsection{Node Configuration Convention}
\label{subsubsection:NodeConfigurationConvention}
The 2D trilateration procedure, commonly used for localization applications, requires the measurement of the distances between the target node, i.e. the node to be localized, and two known-position nodes.
This is not possible in the case of relative cooperative localization, when none of the nodes has a known absolute position, and only the position of the nodes in a relative coordinate frame is desired. 

\begin{figure}[t]
    \centerline{\includegraphics[width=0.9\columnwidth]{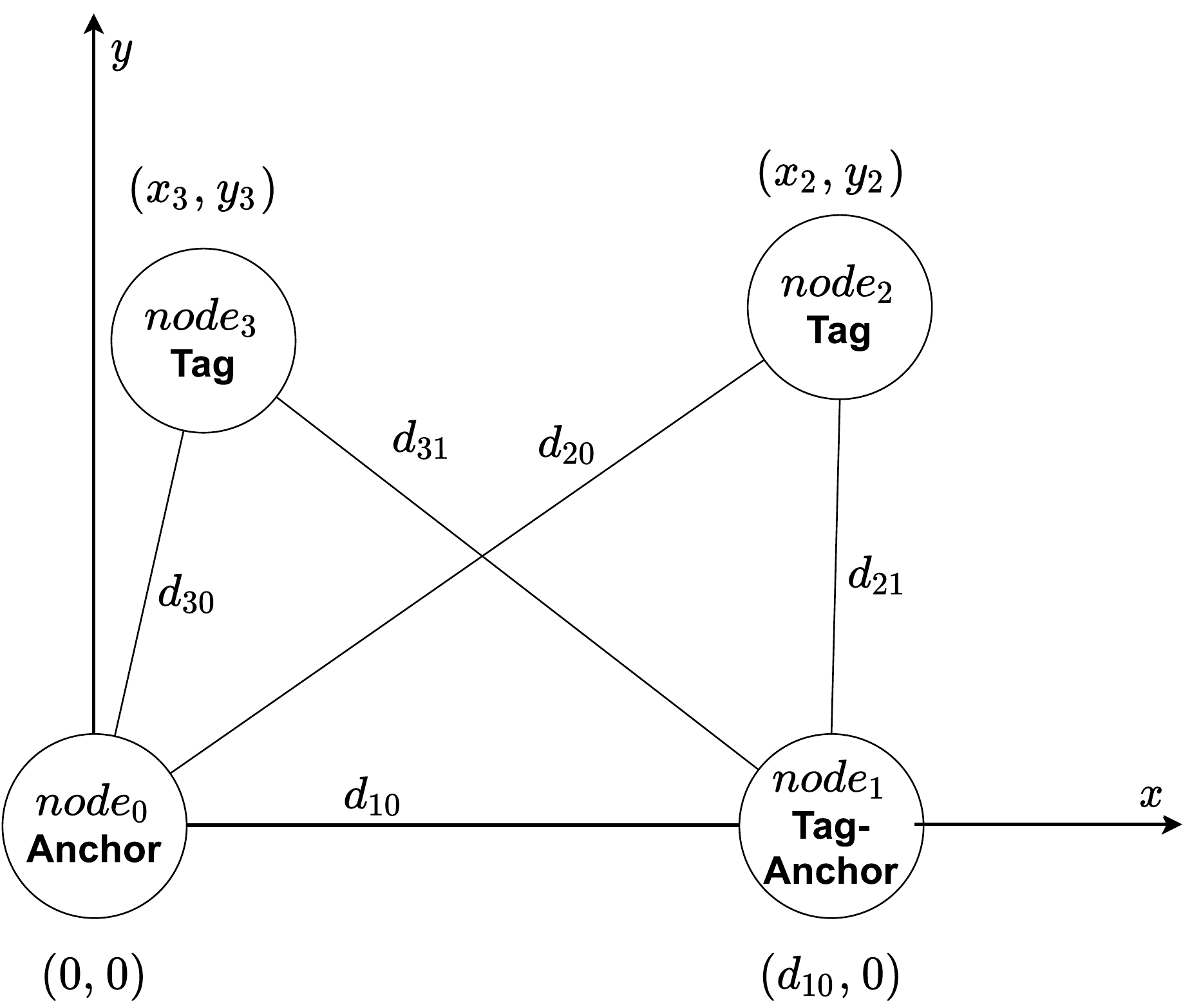}}
    \caption{Node configuration in the relative coordinate frame considered, showing the measured distances.}
    \label{fig:NodeCoordinates}
\end{figure}

To define this relative coordinate frame, a convention is adopted.
Specifically, as illustrated by Fig. \ref{fig:NodeCoordinates}, the following constraints on the $x$ and $y$ coordinates of some of the nodes are imposed:

\begin{itemize}
    \item All nodes are placed on the same plane
    \item $node_{0}$ is the origin of the coordinate frame, i.e. its coordinates are $(0,0)$
    \item $node_{1}$ lies on the $x$ axis. Its $x$ coordinate is the measured distance with respect to $node_{0}$, denoted as $d_{10}$\footnote{Measured distances are denoted as '$d_{xy}$', where x is the tag and y is the anchor}, and its $y$ coordinate is 0, i.e. its coordinates are ($d_{10}$, 0)
    \item $node_{2}$ and $node_{3}$ have no position constraints
\end{itemize}

\subsubsection{Localization Method}
\label{s:localization_method}
Given the convention described in Section \ref{subsubsection:NodeConfigurationConvention}, the positions of $node_{2}$ and $node_{3}$ in the relative coordinate frame are estimated by minimizing the following cost functions, respectively: 

\begin{align}
f_{2}(x_{2},y_{2}) =& (
\sqrt{(x_{2} - x_{0} )^2 + (y_{2} - y_{0} )^2 } -  d_{20})^2 +\notag\\
&(\sqrt{(x_{2} - x_{1} )^2 + (y_{2} - y_{1} )^2 } -  d_{21})^2
\label{eq:CostFunctionNLLSNodo2}
\\\notag\\\notag
f_{3}(x_{3},y_{3}) =& (\notag
\sqrt{(x_{3} - x_{0} )^2 + (y_{3} - y_{0} )^2 } -  d_{30})^2 +\notag\\
&(\sqrt{(x_{3} - x_{1} )^2 + (y_{3} - y_{1} )^2 } -  d_{31})^2
\label{eq:CostFunctionNLLSNodo3}
\end{align}

The cost functions in \eqref{eq:CostFunctionNLLSNodo2} and \eqref{eq:CostFunctionNLLSNodo3} are numerically minimized by the algorithm mentioned in Section \ref{s:software}.
Therefore the $x_n$ and $y_n$ coordinates are obtained as $$\argminA_{x_n,y_n} f_n(x_n,y_n)$$ where $n = 2,3$.

\subsubsection{Calibration}
\begin{figure}[t]
    \centering
    \centerline{\includegraphics[width=\columnwidth]{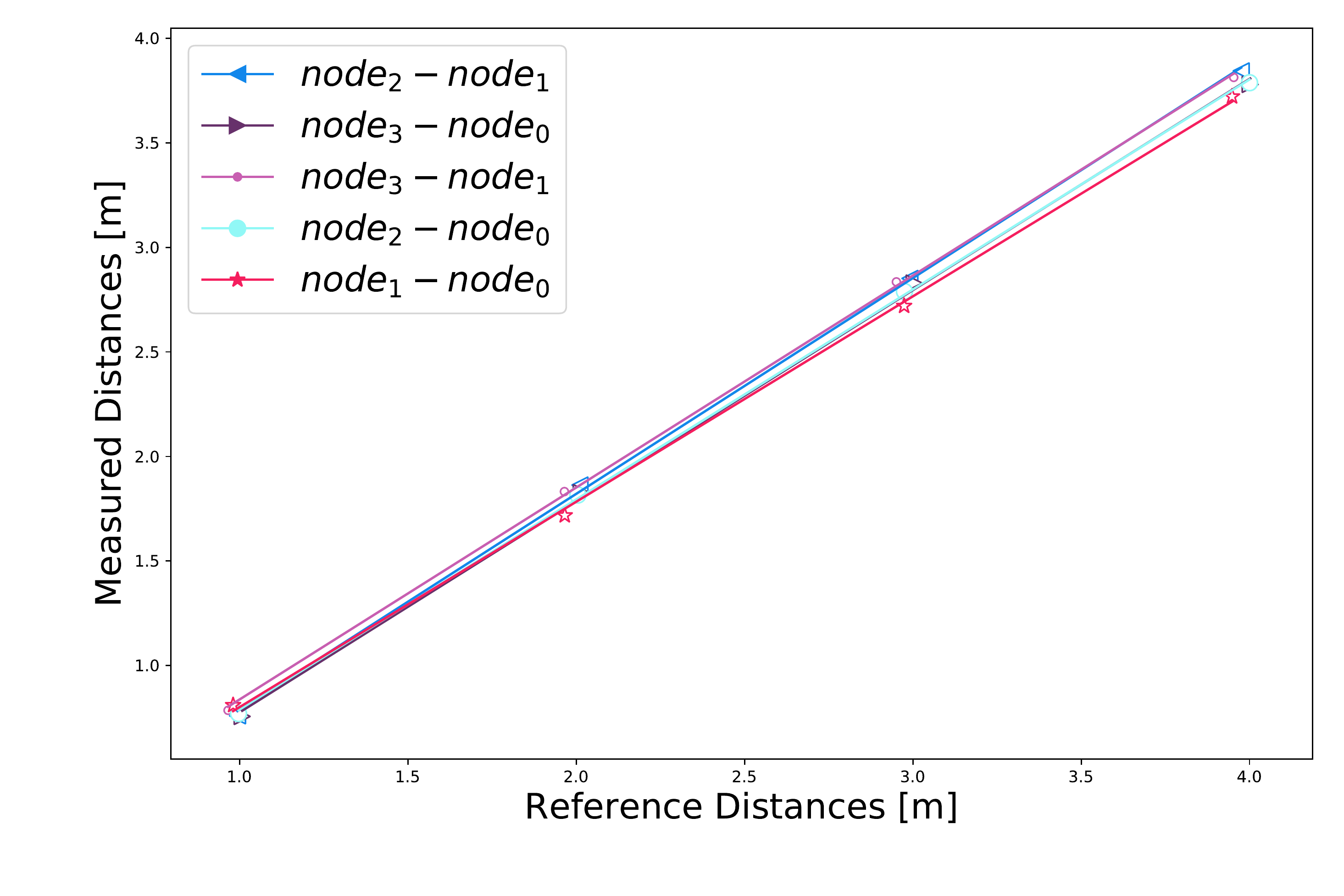}}
    \caption{Calibration curves for each Tag-Anchor couple. The markers denote the experimental data, while the solid lines represent the fitting results.}
    \label{fig:CalibrationCurves}
\end{figure}

\begin{figure}[t]
    \centering
        \subfigure[]{
        \includegraphics[width=0.45\columnwidth]{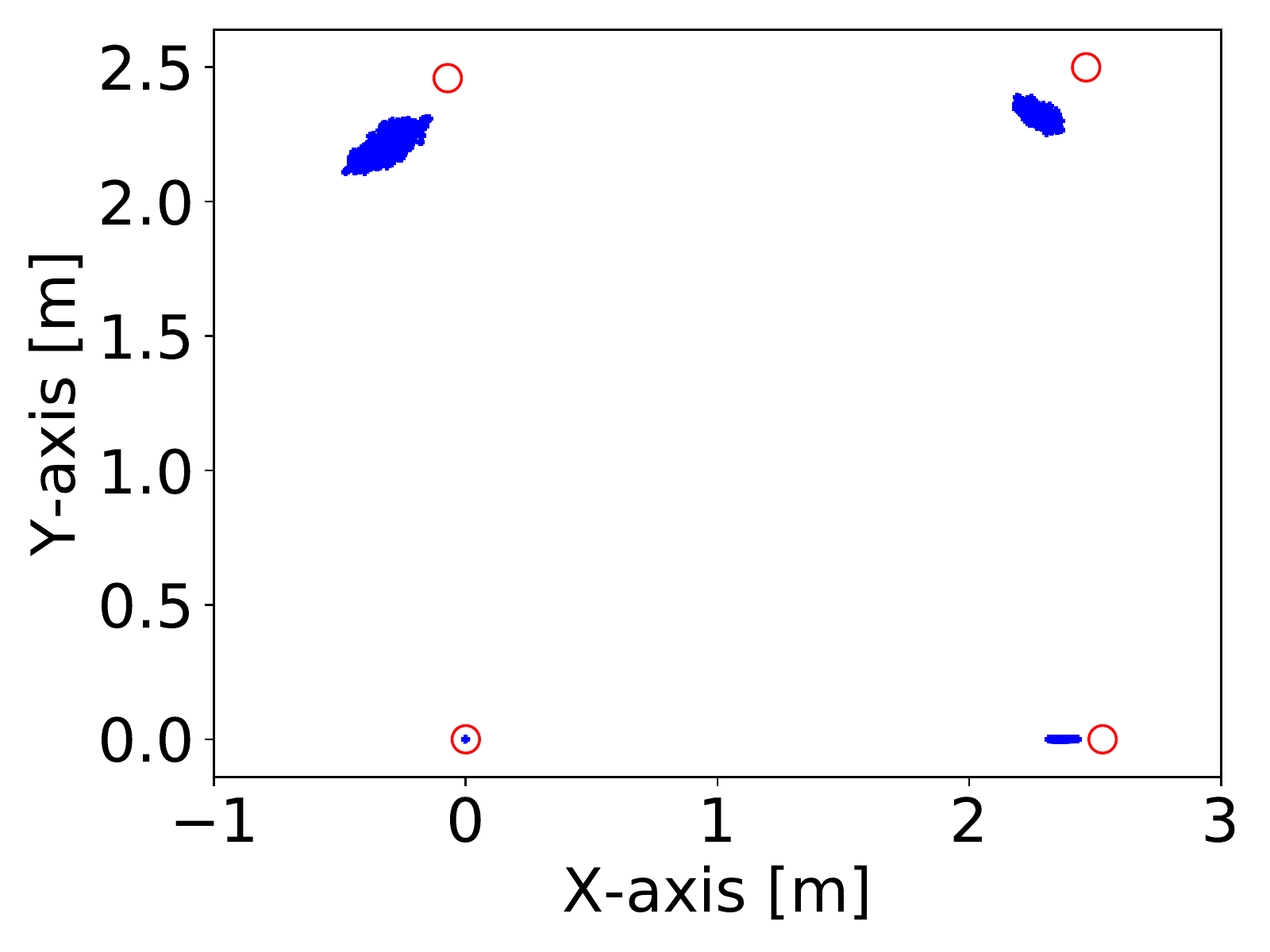}
        \label{fig:pos:square:noCalib}
        }
    \subfigure[]{
        \includegraphics[width=0.45\columnwidth]{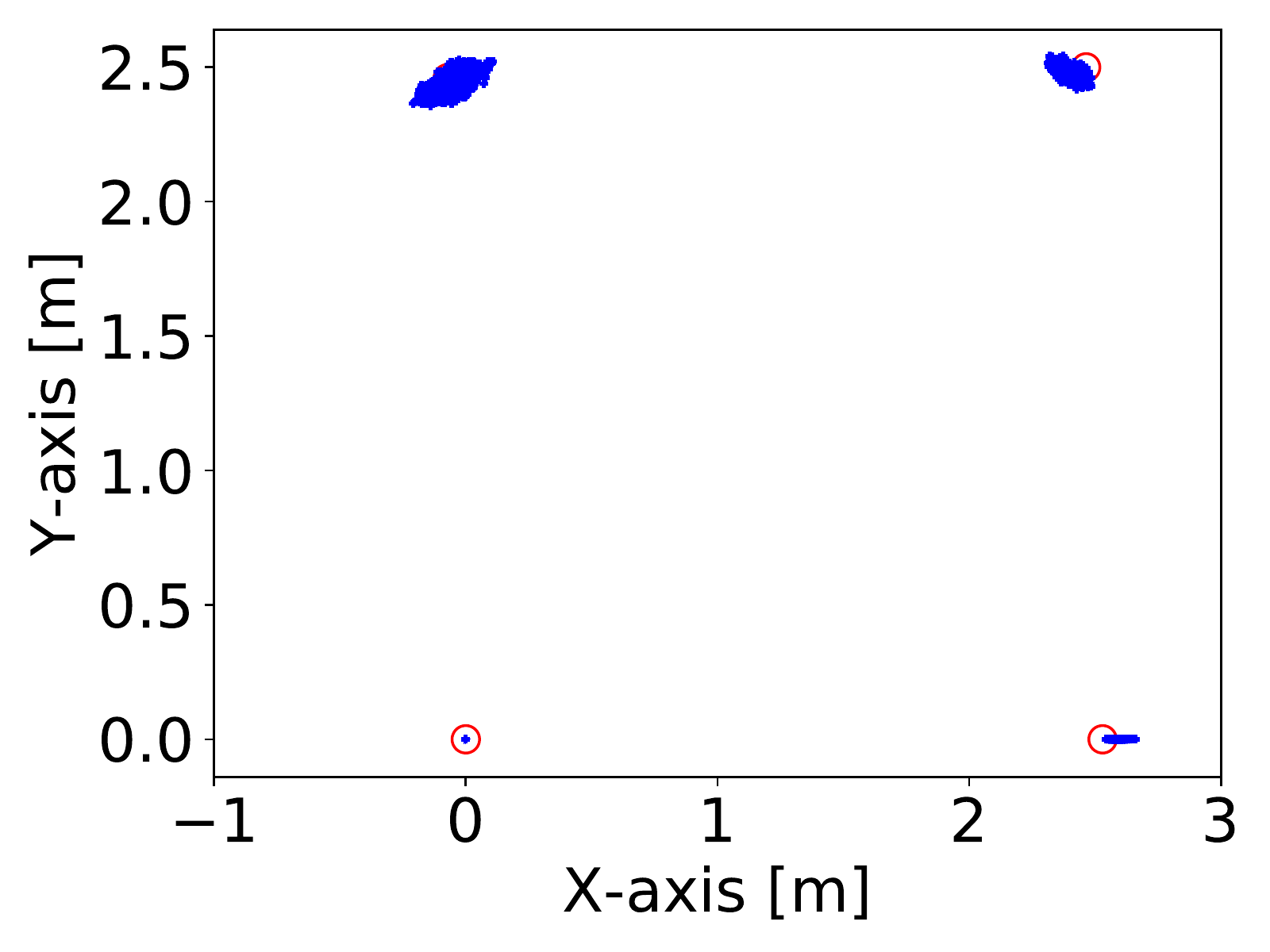}
        \label{fig:pos:square:Calib}
        }
    \caption{Comparison of estimated position without calibration \subref{fig:pos:square:noCalib} and with calibration \subref{fig:pos:square:Calib}, in square configuration. The reference positions are denoted by red circles, and the estimated positions by blue dots. }
    \label{fig:calib_vs_no_calib}
\end{figure}

\begin{figure}[t]
    \centering
        \includegraphics[width=\columnwidth]{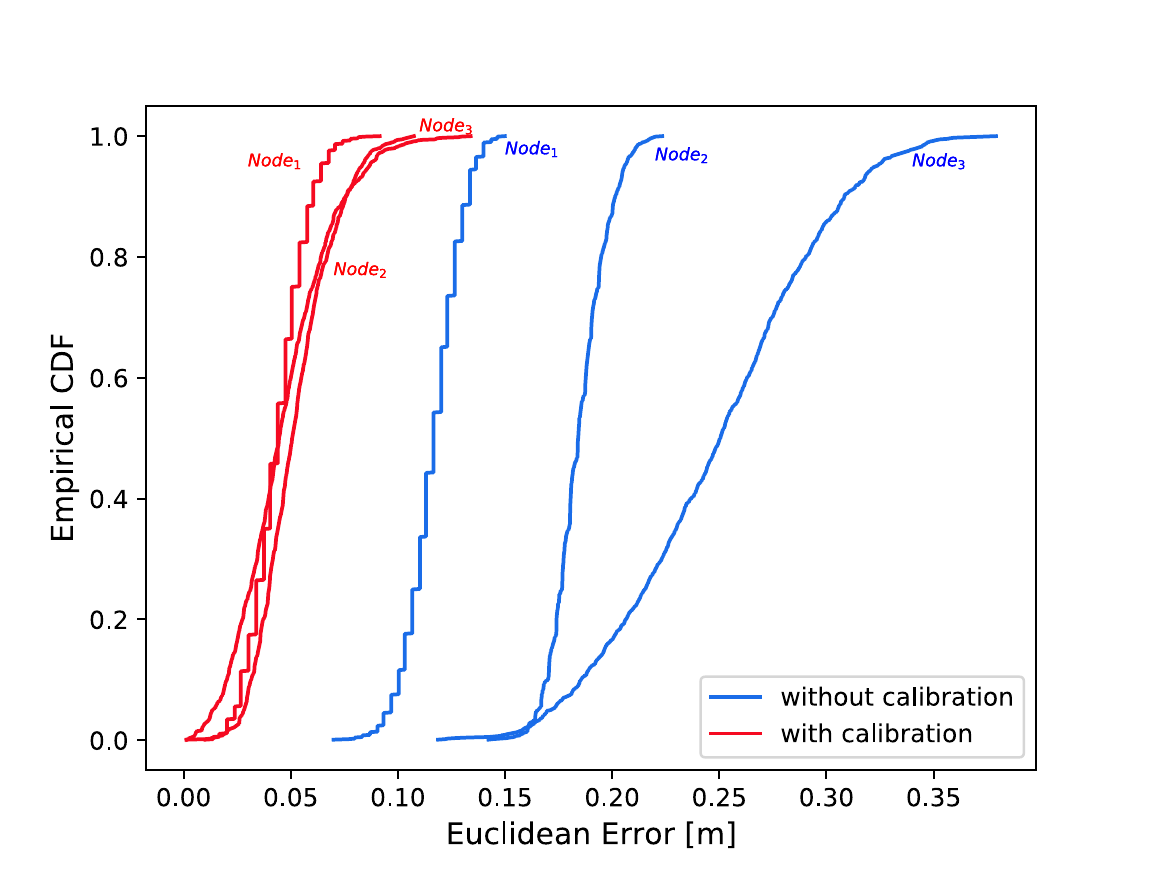}
    \caption{Cumulative Distribution Function of nodes 1,2,3 in square configuration with calibration correction and without it.}
    \label{fig:CDFEuclideanErrorCalibrationVsNoCalibration}
\end{figure}

A calibration procedure, similar to \cite{9459845}, was performed to reduce the effect of the systematic error. 
To perform this procedure, each anchor-tag couple was placed at fixed and known distances, from 1 m to 4 m with 1 m steps.
For each of these configurations, 1200 distance measurements were acquired and averaged.
These measurements were executed in the same external conditions of the experimental tests. 
Fig. \ref{fig:CalibrationCurves} shows the calibration curves from each anchor-tag couple: on the $x$-axis there are the reference distances manually measured with a laser distance meter, while the $y$-axis reports the average distances retrieved by the sensors. From the calibration curves a mathematical linear model is assumed and formulated as follows:
\begin{equation}
d_{m}	\approx m_{c}d_{r}+q_{c}
\label{eq:calibration_mathematical_linear_model}
\end{equation}
where $d_{m}$ is the measured distance, $m_{c}$ the calibration slope, $d_{r}$ the laser reference measurement and $q_{c}$ the calibration intercept.
The parameters $m_c$ and $q_{c}$ are computed by fitting a line to the experimental data. 
Then, the calibrated distances $d_{c}$, are calculated using the calibration equation defined in \eqref{eq:calibration_equation}. 
\begin{equation}
d_{c} = \frac{d_{m}-q_{c}} {m_{c}}
\label{eq:calibration_equation}
\end{equation}
$d_{c}$ is the value used within the position algorithm to estimate the node pose.
The improvements due to the calibrations are highlighted in Fig. \ref{fig:calib_vs_no_calib} wherein an uncalibrated square configuration position estimation is compared with the calibrated one. Moreover, this difference is evident in the cumulative distribution function reported in Fig. \ref{fig:CDFEuclideanErrorCalibrationVsNoCalibration} for nodes 1,2,3 in square configuration: the calibrated nodes show a considerably lower empirical error.

\section{Experimental Results}\label{AA}
\begin{figure}[t]
    \centering
    \includegraphics[width=0.9\columnwidth]{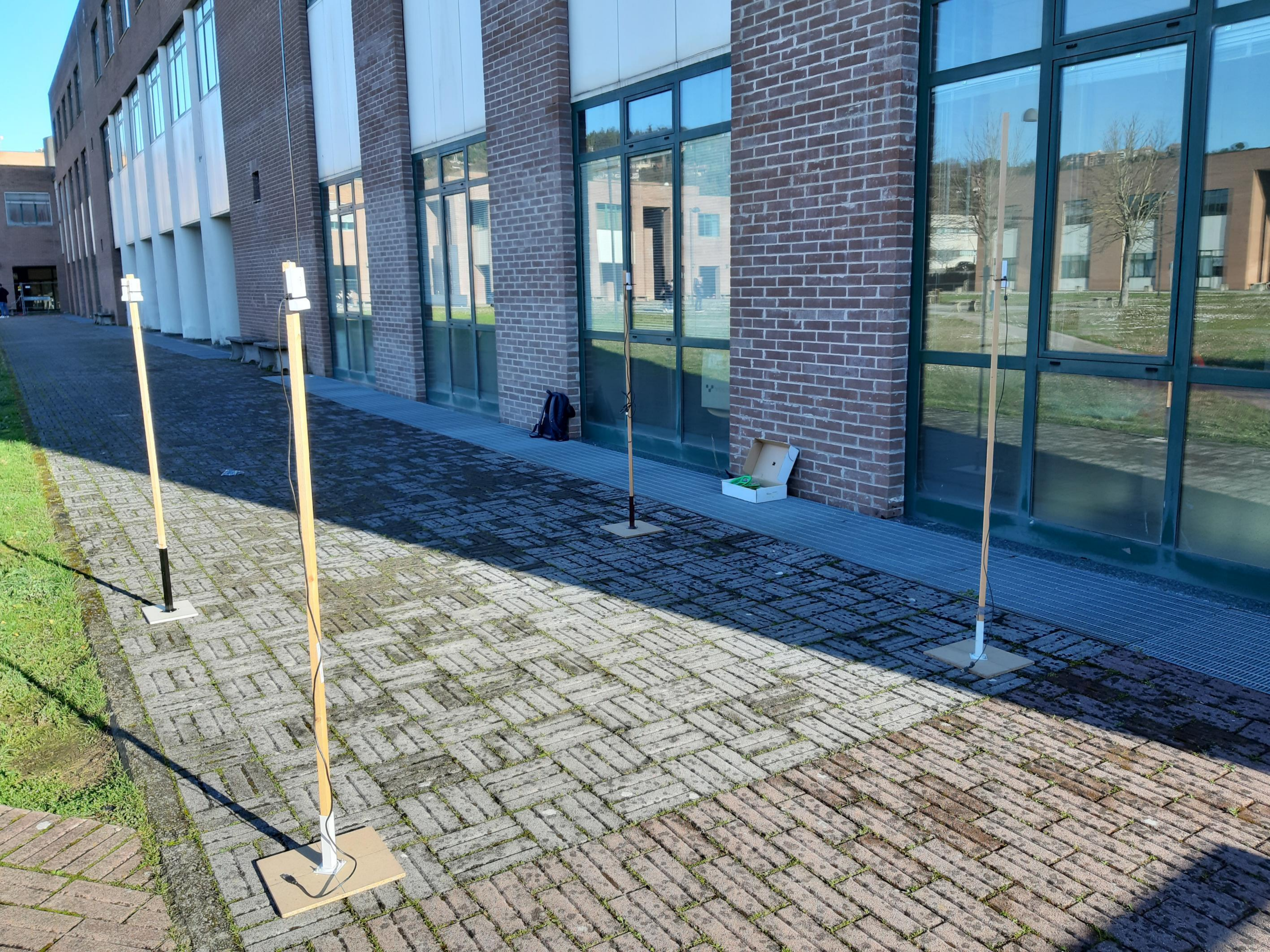}
    \caption{Photo of the experimental setup. The four UWB nodes are mounted on stands and powered by battery packs.}
    \label{fig:photo_experiment}
\end{figure}

To evaluate the position estimation accuracy achieved by the system, experimental tests were performed. 
During these tests, the four nodes were placed on stands in an outdoor environment with line-of-sight conditions, powered by battery packs, as shown in Fig. \ref{fig:photo_experiment}. 
The raw UWB ranging data was acquired via BLE connection.

The Euclidean error \eqref{eq:EuclideanError} was used as the error metric, defined as follows:
\begin{equation}
||\epsilon|| = \sqrt{(\hat{x}-x_t)^2+(\hat{y}-y_t)^2}
\label{eq:EuclideanError}
\end{equation}
where $(\hat{x},\hat{y})$ represent the estimated coordinates and $(x_t,y_t)$ the reference coordinates, which were acquired through manual measurements using a tape measure.
The "Evaluation Scripts" software module in Figure \ref{fig:ProposedSolutionSwStack} was coded to process the results: it calculates the Euclidean error and produces multiple plots from the data read from the Rosbag\footnote{Collection of data organized by topic, provided by ROS.} of each conducted experiment.

\begin{figure*}[tb]
    \centering
    \subfigure[]{
        \includegraphics[width=0.6\columnwidth]{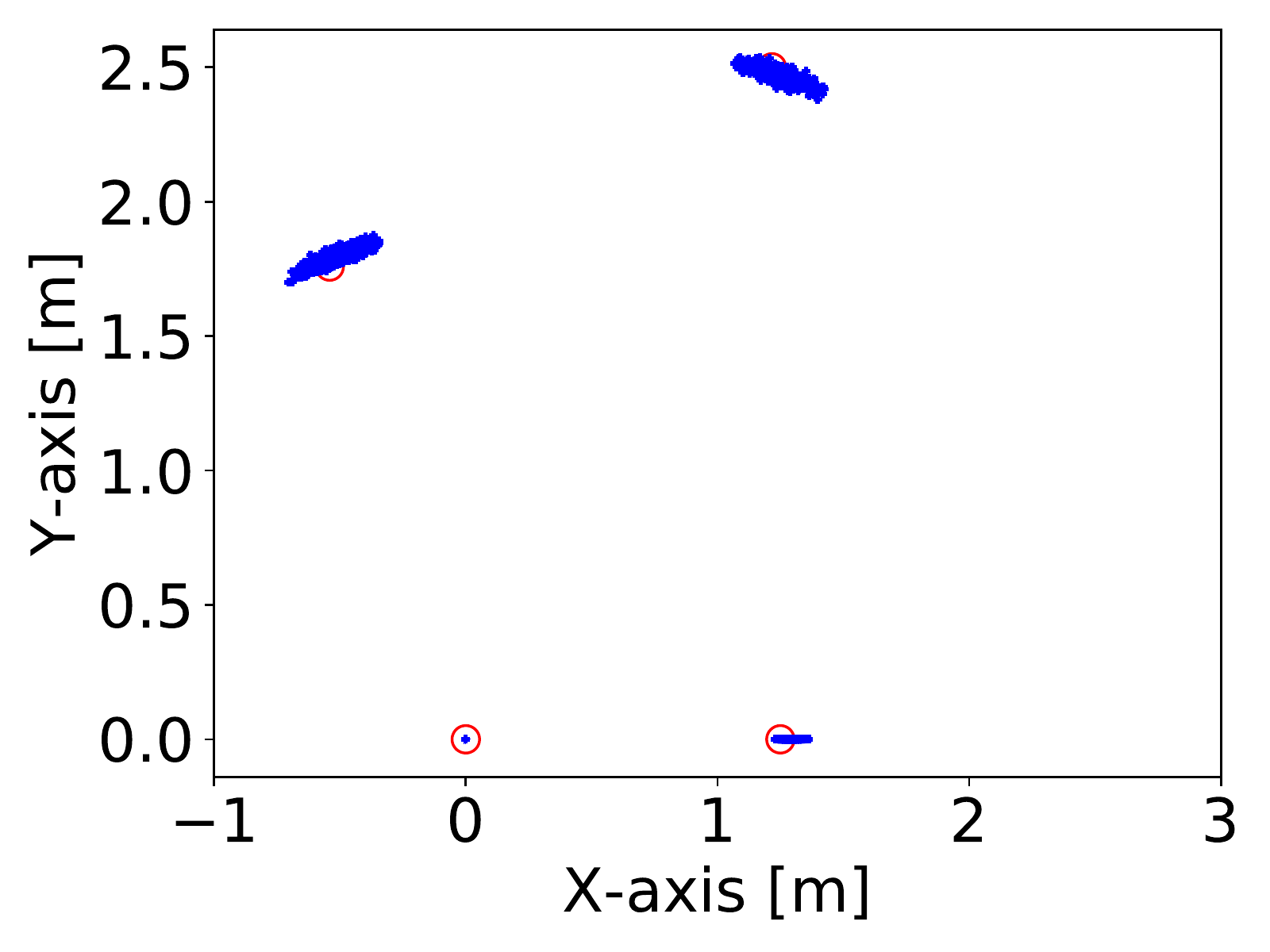}
        \label{fig:pos:quad}
        }
    \subfigure[]{
        \includegraphics[width=0.6\columnwidth]{Figures/Estimate_Positions_for_shape_square_shape_calibrated.pdf}
        \label{fig:pos:square}
        }
    \subfigure[]{
        \includegraphics[width=0.6\columnwidth]{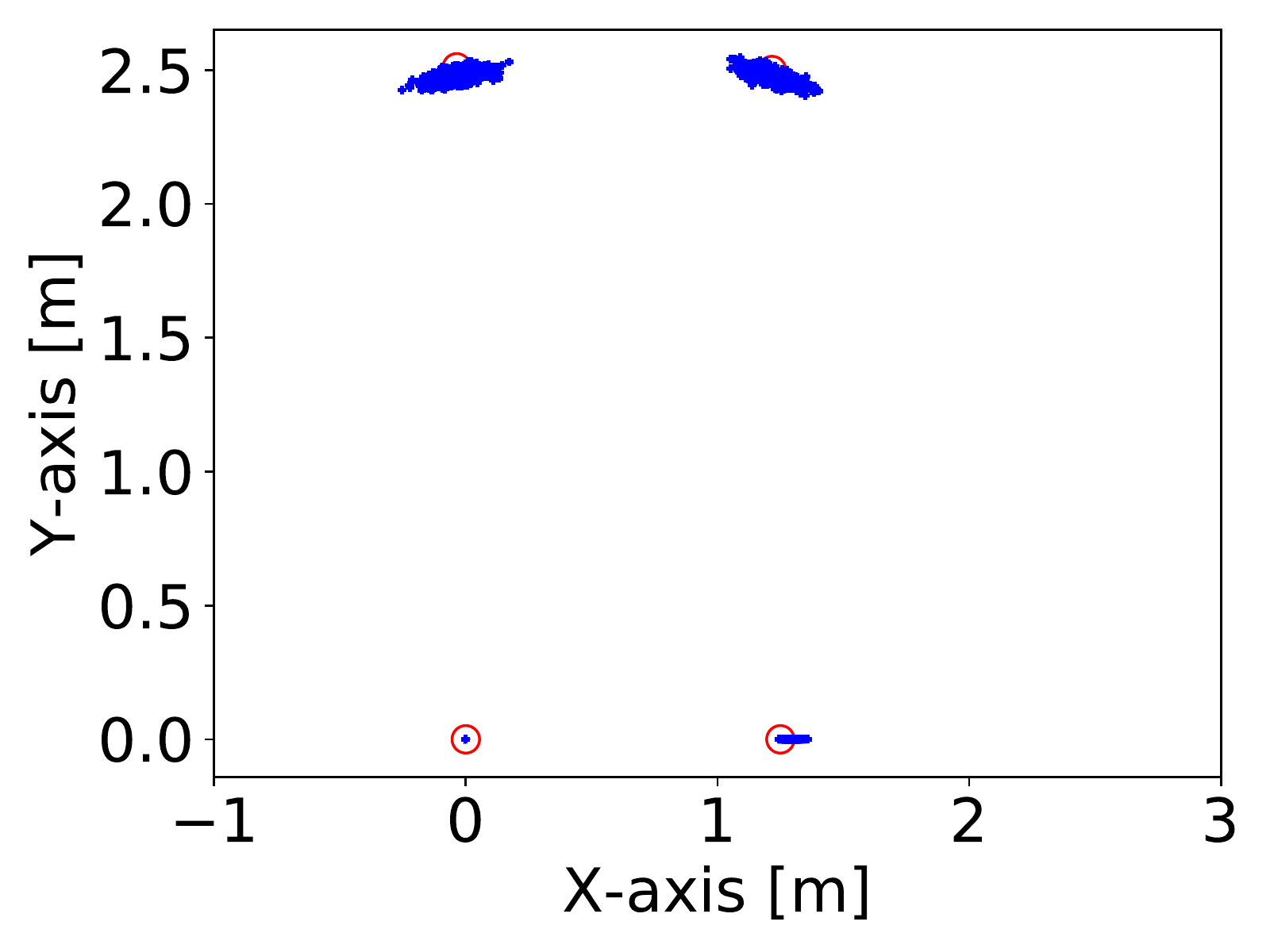}
        \label{fig:pos:rectangular}
        }
        \\ 
        %Box Plots 
    \subfigure[]{
        \includegraphics[width=0.6\columnwidth]{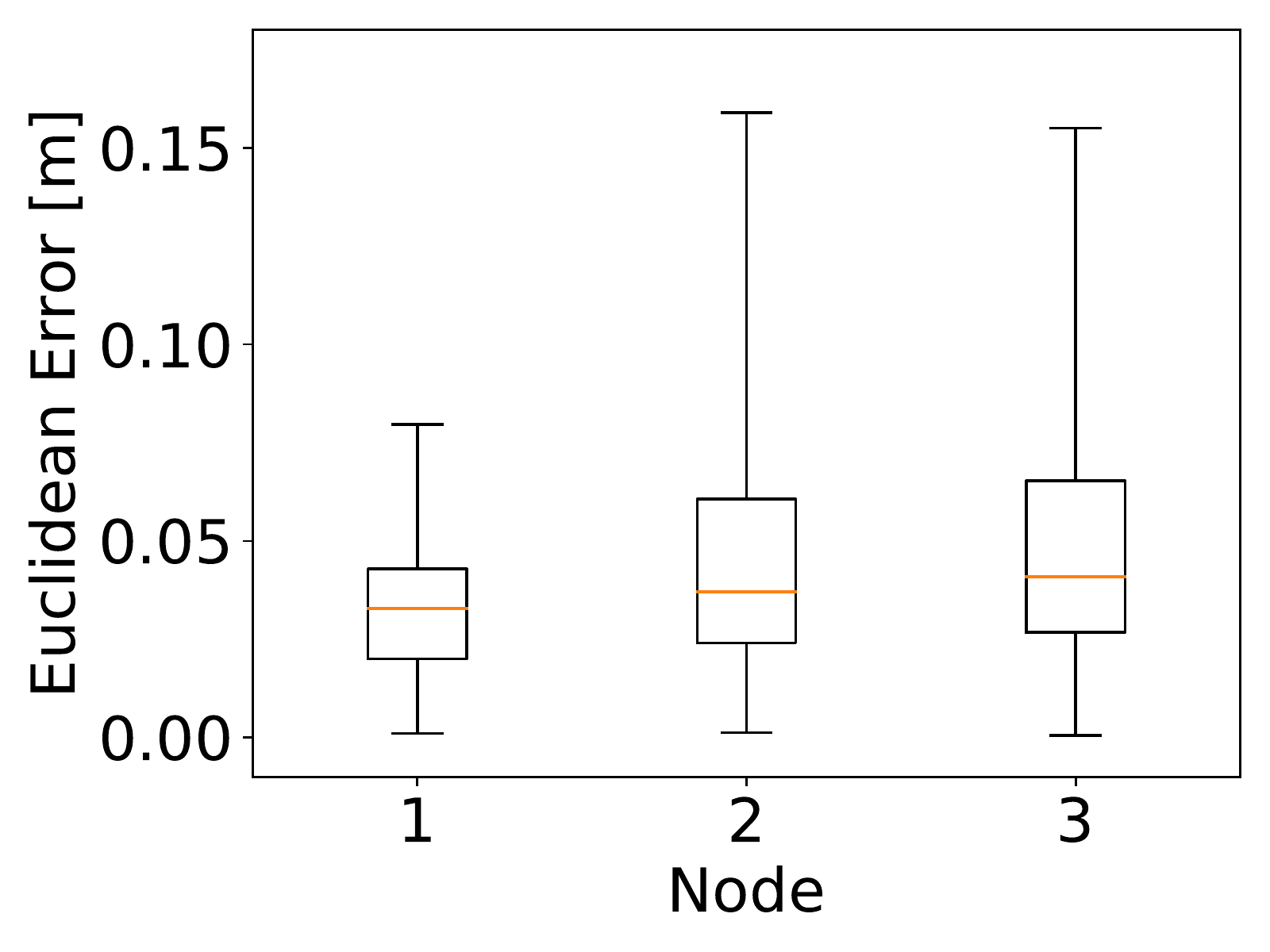}
        \label{fig:box:quad}
        }
    \subfigure[]{
        \includegraphics[width=0.6\columnwidth]{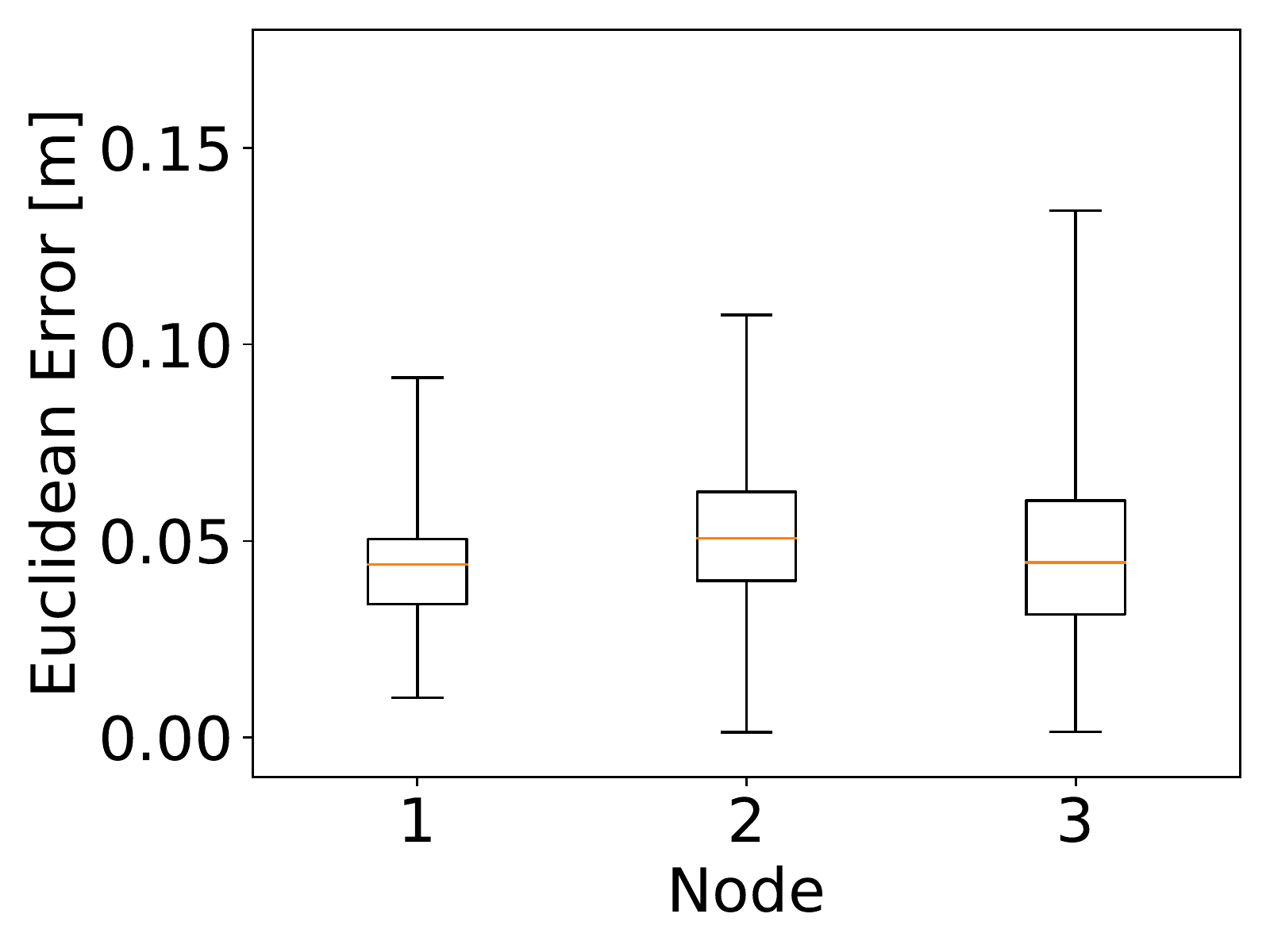}
        \label{fig:box:square}
        }
    \subfigure[]{
        \includegraphics[width=0.6\columnwidth]{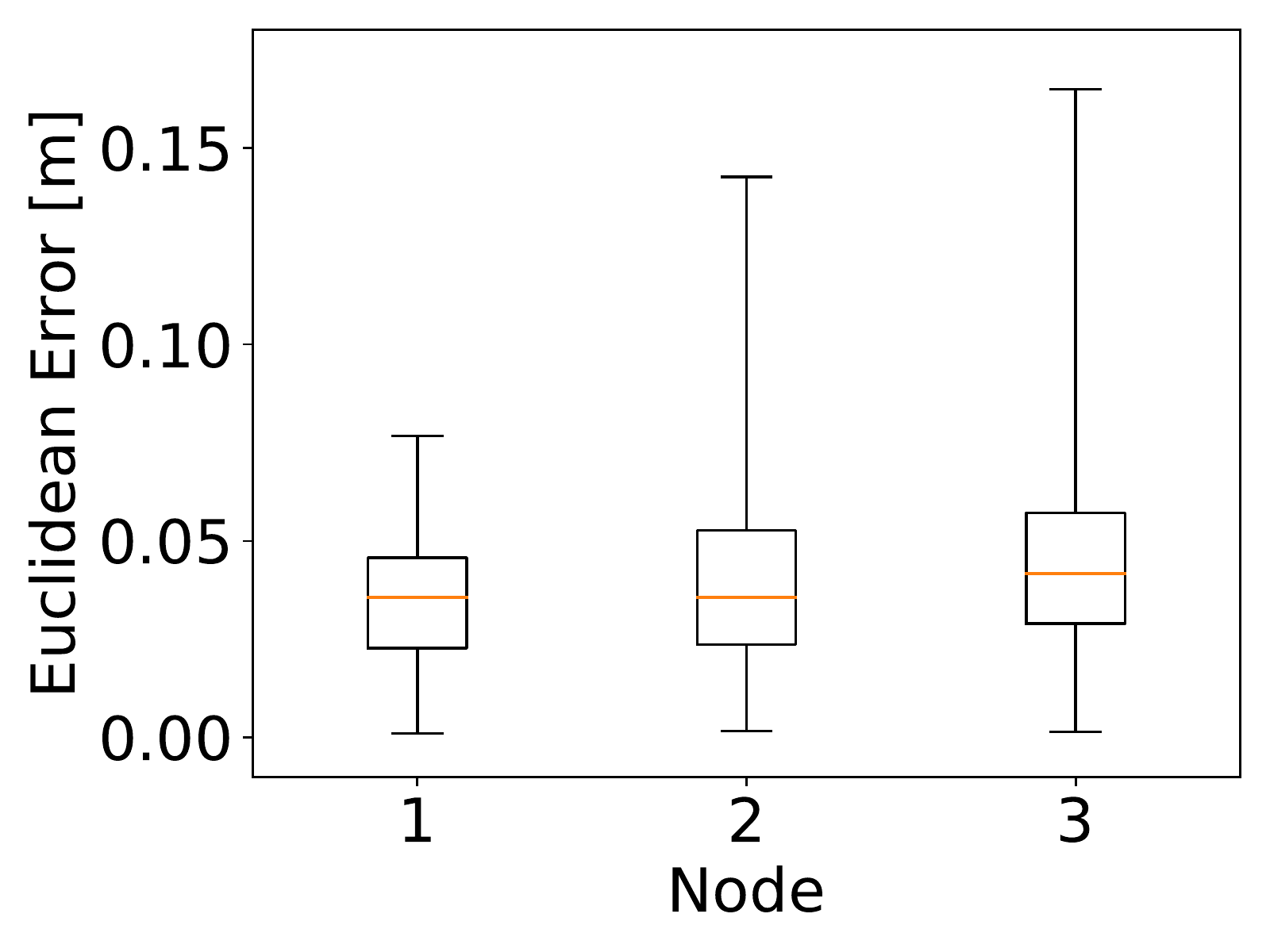}
        \label{fig:box:rectangular}
        }
        \caption{Experimental results and Euclidean error Box-Plots: quadrilateral in \subref{fig:pos:quad} and \subref{fig:box:quad}, square in \subref{fig:pos:square} and \subref{fig:box:square} and  rectangular in \subref{fig:pos:rectangular} and \subref{fig:box:rectangular}}
\end{figure*}

The setups used to carry out the tests are 3 and they only differ in the arrangement of the nodes. Performing the measurements outdoors in an open space, therefore without considerable obstacles capable of generating reflections or attenuation of the signal, conditions have been chosen that approximate the ideal ones. The analyzed node arrangements, instead, were chosen with the intention of stressing the system in critical configurations. Following are the chosen geometries:
\begin{itemize}
        \item square
        \item rectangle
        \item quadrilateral
\end{itemize}
Each different node arrangement was tested for 2 minutes, equivalent to 1200 measurements per node\footnote{the DWM1001 nodes are able to make 10 measurements every second, i.e. with a rate of $10$ Hz}.
The actual configurations and their respective position estimates made through the software architecture described in Section \ref{section:Proposed System}, are displayed in Figures \ref{fig:pos:quad}, \ref{fig:pos:square} and \ref{fig:pos:rectangular}.

From these figures, it is possible to notice that the error in the lower left node ($node_{0}$) is always null.
Moreover, the position estimates of the lower right node ($node_{1}$) vary only with respect to the $x$ component of the coordinates. 
This reflects the convention adopted for relative localization, described in Section \ref{subsubsection:NodeConfigurationConvention}.
Finally, the other two nodes have position estimates that vary on both axes.

\begin{figure}[t] 
    \centerline{\includegraphics[width=\columnwidth]{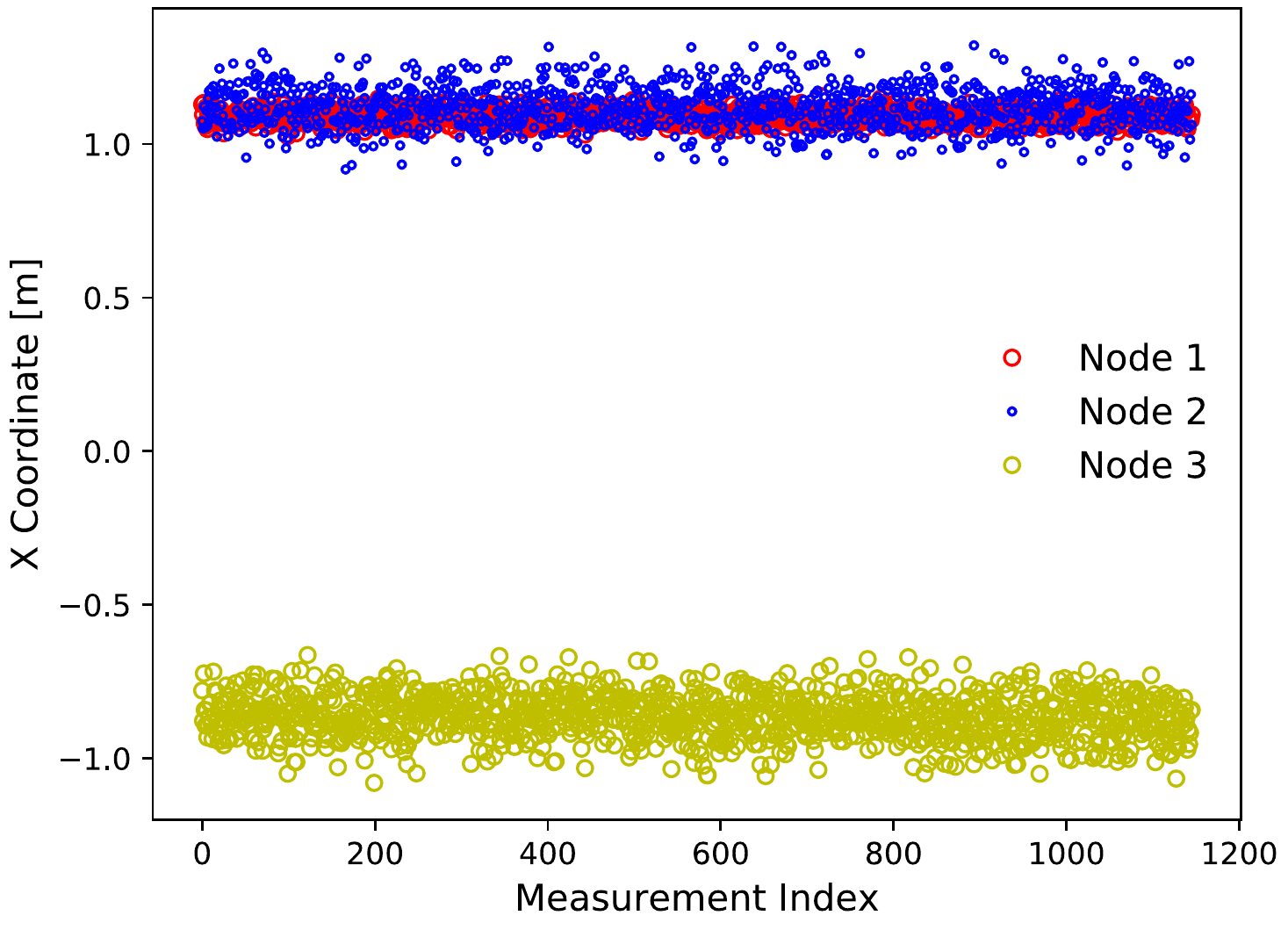}}
    \caption{Quadrilateral configuration estimated $x$ coordinates.}
    \label{fig:QuadrilateralConfigurationsemi_generic_quadrilateral_shapeXAxisCoordinatesEstimatesPlt}
\end{figure}

The distribution of the Euclidean error is shown by the box-plots of Figures \ref{fig:box:quad}, \ref{fig:box:rectangular} and \ref{fig:box:square}, for each configuration.
From these figures, it can be observed that the maximum error is always below 17 cm.
Furthermore, the graph of the estimates of the $x$ coordinate in the quadrilateral configuration is shown in Figure \ref{fig:QuadrilateralConfigurationsemi_generic_quadrilateral_shapeXAxisCoordinatesEstimatesPlt}, which highlights how there are no measurement outliers.

Finally, Table \ref{table:MeanSquareError} provides a comparison between the RMSE (Root Mean Square Error) values of all the nodes for each test performed.
Note how Node 0, being the origin of the chosen reference system, always has a null error. 
There are no remarkable error differences between nodes and configurations.
All errors are, however, less than, on average, 5 cm for every node and 4 cm for every geometrical configuration.
Finally, it can be stated that the average error in the position estimation obtained by the proposed system is 3.2 cm.
Our 2D-10Hz cooperative localization system shows comparable results with the 3D-1Hz system described in \cite{9341042}.
%%%%%%%%%%%%%%%%%%%%%%% NON CALIBRATED DATA %%%%%%%%%%%%%%%%%%%%%%%%%%%%%%%%%%%%
\begin{comment}
    \begin{table}[htbp]
    \centering
    \caption{RMSE, in meters, grouped by configuration and node}
    \label{table:MeanSquareError}
    \begin{threeparttable}
    \setlength\tabcolsep{0pt}
    \begin{center}
    \begin{tabular*}{\columnwidth}{@{\extracolsep{\fill}} lllll ccccc}
    \hline
                             & \multicolumn{4}{c}{Node}          &           \\
    Configuration                    & Node 0 & Node 1 & Node 2 & Node 3 & Node Mean \\ \cline{2-6} 
    \addlinespace
    Square                   & 0      & 0.116  & 0.185  & 0.249  & 0.137     \\
    Rectangular              & 0      & 0.110  & 0.166  & 0.305  & 0.145     \\
    Quadrilateral & 0      & 0.113  & 0.161  & 0.340  & 0.153     \\
    \addlinespace
    Configuration Mean               & 0      & 0.113  & 0.170  & 0.298  & 0.145    
    \end{tabular*}
    \end{center}
    \end{threeparttable}
    \end{table}
\end{comment}
%%%%%%%%%%%%%%%%%%%%%%% NON CALIBRATED DATA %%%%%%%%%%%%%%%%%%%%%%%%%%%%%%%%%%%%

%%%%%%%%%%%%%%%%%%%%%%% CALIBRATED DATA %%%%%%%%%%%%%%%%%%%%%%%%%%%%%%%%%%%%
\begin{table}[htbp]
\centering
\caption{RMSE, in meters, grouped by Configuration and node}
\label{table:MeanSquareError}

\setlength\tabcolsep{0pt}
\begin{center}
\begin{tabular*}{\columnwidth}{@{\extracolsep{\fill}} lllll ccccc}
\hline
                         & \multicolumn{4}{c}{Node}          &           \\
Configuration                    & Node 0 & Node 1 & Node 2 & Node 3 & Node Mean \\ \cline{2-6} 
\addlinespace
Square                   & 0      & 0.044  & 0.052  & 0.047  & 0.036     \\
Rectangular              & 0      & 0.035  & 0.040  & 0.045  & 0.030     \\
Quadrilateral & 0      & 0.032  & 0.045  & 0.048  & 0.031     \\
\addlinespace
Configuration Mean               & 0      & 0.037  & 0.046  & 0.047  & 0.032    
\end{tabular*}
\end{center}

\end{table}
%%%%%%%%%%%%%%%%%%%%%%% CALIBRATED DATA %%%%%%%%%%%%%%%%%%%%%%%%%%%%%%%%%%%%

\section*{Conclusions}
In this work, a cooperative localization system based on UWB technology has been presented. 
The developed system presents a fully-wireless setup, in which ranging data are acquired using BLE connections, and performs relative positioning of four nodes. 
The APIs provided by the devices used to obtain real-time data have been exploited.
The minimization problem associated with the relative localization, described by a cost function with the constraints imposed by our convention, is solved through a non-linear least squares algorithm. The results show good performance, with an RMSE of 3.2 cm, which is of the same order of magnitude as DWM1001's capabilities according to the datasheet.
However, by analyzing the RMSE results in different configurations, we also notice how the geometry of the problem, along with the restricted set of measurements for each node, may negatively affect the performance in extreme cases.
This can potentially be mitigated by exploiting a network in which ranges are measured between all nodes (fully-connected network), obtainable using pairs of sensors in each node of the network. A further step to be explored concerns the analysis of system performance in dynamic situations. As an example, by placing the nodes on top of mobile robots (drones or ground robots) in motion, it could be possible to evaluate whether the accuracy of position estimation is suitable for shape-formation tasks.

\bibliographystyle{IEEEtran}

% \bibliography{bibliografia.bib}

\end{document}